\PassOptionsToPackage{pagenumbers}{wacv}

\documentclass[10pt,twocolumn,letterpaper]{article}

\usepackage{wacv}              

\definecolor{wacvblue}{rgb}{0.21,0.49,0.74}
\usepackage{xr}
\usepackage[pagebackref,breaklinks,colorlinks,allcolors=wacvblue]{hyperref}
\usepackage{multirow}
\usepackage{svg}

\def\wacvPaperID{1359} 
\def\confName{WACV}
\def\confYear{2027}

\title{FutureLift: Lifting Trajectory Proposals into Bidirectional Future Graphs for Motion Forecasting}

\author{
Daniel Grimm \qquad J. Marius Z\"ollner\\
FZI Research Center for Information Technology\\
{\tt\small \{grimm,zoellner\}@fzi.de}
}

\begin{document}
\NewCommandCopy{\combinedcref}{\cref}
\NewCommandCopy{\combinedCref}{\Cref}
\RenewDocumentCommand{\cref}{s m}{\combinedcref{#2}}
\RenewDocumentCommand{\Cref}{s m}{\combinedCref{#2}}

\maketitle
\begin{abstract}
    Accurate motion prediction is critical for safe autonomous driving, yet even strong forecasters leave residual errors in their multimodal proposals. Existing refiners compress complete proposals or update their segments sequentially, preventing the full proposed future from serving as explicit relational context for correction. We introduce \textbf{FutureLift}, a backbone-agnostic refiner that represents each multimodal proposal as a heterogeneous future graph, with spatially grounded temporal nodes connected to agent histories, map elements, alternative modes, and, for joint prediction, other agents future trajectories. Bidirectional temporal edges allow later evidence to revise earlier segments in a single pass, and the updated nodes predict residual coordinate and confidence corrections. Requiring only proposals, scores, and scene context, FutureLift provides a common refinement interface for structurally different forecasting backbones. On Argoverse~2 and the Waymo Open Motion Dataset, it consistently improves frozen proposals and outperforms native and specialized refiners across marginal and joint forecasting. On the WOMD motion-prediction leaderboard, FutureLift ranks first in minADE (0.5049~m) and minFDE (0.9942~m) among the compared standard-input methods, without LiDAR or ensembling, while achieving a competitive 0.4874 Soft mAP.
    Code will be available at \href{https://url.fzi.de/futurelift}{\textcolor[HTML]{EC008C}{url.fzi.de/futurelift}}.
\end{abstract}

\section{Introduction}
\label{sec:intro}

\begin{figure}[t]
    \centering
    \includegraphics[width=\columnwidth]{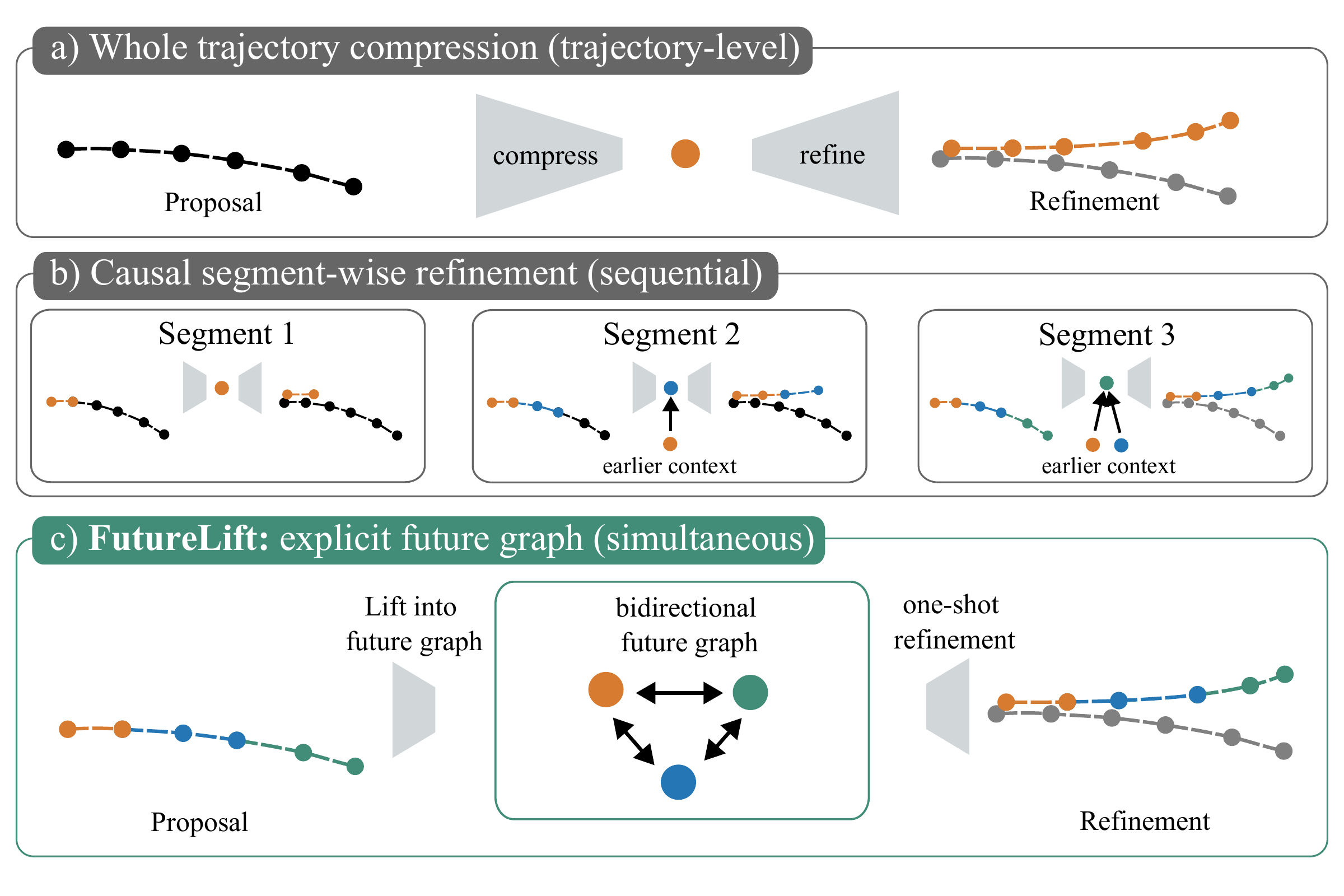}
    \caption{\textbf{Refinement representations.} \textbf{Top:} Trajectory-level methods compress proposal geometry into a shared representation. \textbf{Middle:} Causal segment-wise methods pass information forward, so later evidence cannot reach earlier segments within a forward-only sweep. \textbf{Bottom:} FutureLift represents all proposal segments in an explicit future graph and refines them simultaneously, allowing later context to correct earlier segments through bidirectional temporal relations.}
    \label{fig:refinement_comparison}
\end{figure}

Accurate motion forecasting enables autonomous vehicles to anticipate traffic participants and plan safely. Given agent histories and an HD map, models must represent multiple plausible futures over an extended prediction horizon. Modern methods use vectorized scene representations \cite{lanegcn, vectornet}, graph- or attention-based context encoders \cite{transformer, graphtransformer, holigraph}, and expressive trajectory decoders \cite{qcnet, donut}. Yet even strong predictors leave residual displacement and ranking errors in their multimodal proposals.

Trajectory refinement can correct such errors without redesigning the underlying predictor. Existing refiners encode trajectory features, retrieve context around selected anchors, or model trajectory--trajectory and trajectory--lane relations \cite{smartrefine, srefiner}. However, compressed embeddings, sparse anchors, and pairwise summaries do not spatially instantiate the complete proposed horizon. This is especially limiting for temporal correction: within a causal segment-wise sweep, later evidence cannot revise an earlier segment \cite{smartrefine, donut}. Iterative relational refiners can revisit predictions \cite{srefiner}, but require repeated computation and still reason primarily at the trajectory level. We instead argue that the complete proposed future should be explicit before correction, allowing every segment to interact jointly with the map, agent histories, alternative modes, and other predicted agents while exchanging evidence in both temporal directions.

Based on this insight, we introduce \textbf{FutureLift}, a backbone-agnostic refiner that lifts multimodal proposals into bidirectional heterogeneous future graphs, as illustrated in \cref{fig:refinement_comparison}. FutureLift discretizes each trajectory proposal into spatially grounded temporal nodes connected to agent histories, map elements, alternative modes, and compatible predictions of other agents. This graph represents a hypothetical future scene rather than a compressed trajectory token. A bidirectional graph network refines the complete horizon in one pass and predicts residual coordinate and confidence updates. Requiring only proposal coordinates, scores, and scene context, FutureLift attaches to structurally different backbones without modifying their proposal mechanisms. The same interface supports marginal and coherent joint-world proposals, enabling social future relations only for the latter.

We evaluate FutureLift on Argoverse~2 and the Waymo Open Motion Dataset using multiple backbones in marginal and joint settings, including within-backbone comparisons with native and specialized refiners. These experiments test interface compatibility rather than transferring a learned refiner between backbones or datasets: each refiner is trained for its target setting. FutureLift consistently improves the underlying proposals; on the WOMD test set, it achieves 0.5049 minADE and 0.9942 minFDE, the best results among the compared standard-input methods, while maintaining competitive Soft mAP. A separate horizon-resolved analysis on the WOMD validation split shows that backward temporal information flow becomes increasingly beneficial at longer horizons.

In summary, our main contributions are:
\begin{itemize}
    \item \textbf{Proposal-to-Future-Graph Lifting:} A backbone-agnostic interface transforms multimodal proposals into heterogeneous future graphs for temporal, map, modal, and joint-only social reasoning.
    \item \textbf{One-Shot Bidirectional Refinement:} One fixed-graph pass lets later future context correct earlier errors without repeated causal sweeps.
    \item \textbf{Interface Compatibility and Temporal-Direction Analysis:} Experiments with multiple backbones on two datasets and in marginal and joint tasks establish consistent gains after setting-specific training, while horizon-resolved validation analysis measures how the contribution of backward temporal information changes with prediction horizon.
\end{itemize}

\section{Related Work}
\label{sec:related_work}

\subsection{Structured Representations for Motion Forecasting}
Vectorized representations preserve agent and map geometry without the quantization and computational cost of rasterized scene encodings. VectorNet~\cite{vectornet} represents scene elements as polylines, while LaneGCN~\cite{lanegcn} constructs a lane graph with explicit actor--map interactions. Later approaches capture richer relational structure: HDGT~\cite{hdgt} introduces heterogeneous node and edge types, GoRela~\cite{gorela} uses viewpoint-invariant relative representations, and HoliGraph~\cite{holigraph} retains fine-grained spatial and temporal entities in a heterogeneous scene graph. These representations primarily organize the observed scene from which future trajectories are decoded.

Recent architectures increasingly place relational reasoning in the decoder. Query-centric methods such as MTR~\cite{mtr}, QCNet~\cite{qcnet}, and DeMo~\cite{demo} progressively decode multimodal futures; FutureNet~\cite{futurenet} incorporates predicted future context into joint forecasting; and DONUT~\cite{donut} formulates trajectory generation as decoder-only autoregressive modeling. Foresight in Motion~\cite{foresightmotion} applies bidirectional state-space modeling to trajectory-state queries within a hierarchical end-to-end decoder, while NDPNet~\cite{ndpnet} separates near- and distant-horizon decoding and promotes kinematically coherent transitions. In these methods, the future representation is part of a specific forecasting architecture. FutureLift instead operates after proposal generation: it lifts the temporal segments of already generated proposals into a common graph for backbone-independent refinement.

\subsection{Trajectory Refinement}
Refinement is often integrated into the forecasting decoder. MTR~\cite{mtr}, for example, repeatedly updates intention queries using local motion information, while QCNet~\cite{qcnet} follows proposal generation with a dedicated refinement stage. Such refiners can exploit backbone features directly, but this coupling makes them difficult to transfer across proposal generators.

Post-hoc methods provide a clearer interface between proposal generation and correction. SmartRefine~\cite{smartrefine} uses scenario-adaptive anchors and local context to recurrently refine trajectory segments, with optional outer iterations for difficult cases. Within a forward sweep, information propagates only from earlier to later segments, while outer iterations repeat the causal refinement process. SRefiner~\cite{srefiner} targets multi-agent refinement by constructing a soft-braid topology over trajectory--trajectory and trajectory--lane relations and iteratively updating joint predictions. Although its paper formulates the initial trajectory representation as an encoding of proposal geometry and agent pose, the released AV2 implementation additionally consumes a precomputed 134-dimensional per-mode embedding exported with each proposal. SRefiner is therefore the closest prior work in combining post-hoc joint refinement with an explicit relational structure, but its released interface is not strictly proposal-only.

FutureLift differs in both representation and interface. Every temporal segment of each multimodal proposal becomes a graph node, and bidirectional temporal edges expose the complete horizon during a refinement pass. The refiner reconstructs its scene representation from proposal coordinates, confidence scores, and original scene inputs; it does not require the proposal backbone's exported embeddings, queries, or intermediate activations.

\subsection{Marginal and Joint Motion Forecasting}
Marginal forecasting predicts each agent's future distribution independently, whereas joint forecasting represents scene-level hypotheses whose agent trajectories are mutually consistent. Many high-performing forecasters focus on marginal prediction~\cite{qcnet, demo}, while FJMP~\cite{fjmp}, FutureNet~\cite{futurenet} explicitly model multi-agent futures. This distinction constrains relational refinement: equally indexed modes from independent marginal predictions need not describe the same joint world, and connecting them can therefore introduce artificial dependencies.

FutureLift preserves this distinction through conditional graph construction. Marginal proposals are refined without cross-agent mode correspondence, whereas scene-consistent joint hypotheses additionally receive social future edges. This design supports both settings without imposing joint semantics on marginal predictions.

\begin{figure*}[t]
    \centering
    \includegraphics[width=\textwidth]{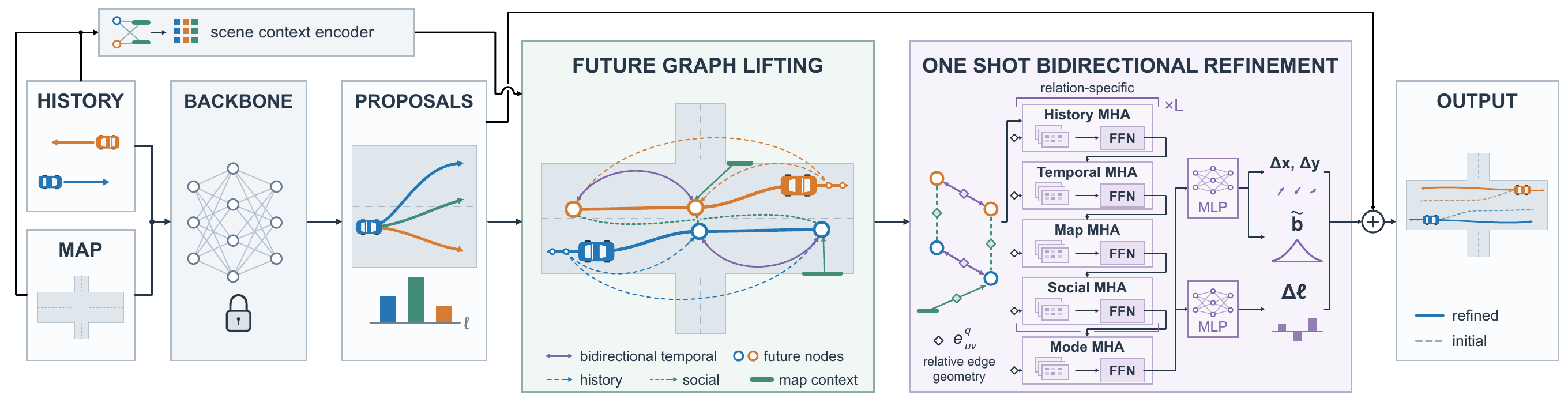}
    \caption{\textbf{FutureLift pipeline.} A frozen backbone supplies trajectory proposals and logits; a separate encoder extracts history and map context without reusing backbone features. Lifting anchors future nodes at proposal-segment endpoints and adds history, bidirectional temporal, map, and social relations. One joint mode is shown schematically; social relations are disabled for marginal proposals. Refinement sequentially applies history, temporal, map, and social attention for $L$ layers, followed by cross-mode attention. Each relation uses its own FFN and relative edge geometry (diamonds). MLP heads decode all intervals on the fixed graph, predicting additive coordinate and logit corrections and Laplace uncertainty scales $\widetilde{\mathbf{b}}$. Solid and dashed output curves denote refined and initial trajectories, respectively.}
    \label{fig:futurelift_overview}
\end{figure*}

\section{Method}
\label{sec:method}
FutureLift is a post-hoc refinement module that turns trajectory proposals into an explicit hypothesis of the future scene. Its central operation is \emph{lifting}: rather than compressing a complete proposal into one latent token, we instantiate its temporal segments as spatially anchored graph nodes. This representation makes the proposal geometry available to map, history, modal, and social reasoning. A single bidirectional graph update then predicts residual corrections to the proposal coordinates and confidence scores. We first formalize the proposal interface, then describe proposal-to-graph lifting, one-shot refinement, and the training objective.

\subsection{Problem Formulation and Proposal Interface}
\label{subsec:interface}

Let a scene contain observed agent histories $\mathcal{H}$, an HD map $\mathcal{M}$, and $A$ agents to be predicted over $T$ future steps. A proposal backbone $B$ produces $K$ candidate trajectories and their logits,
\begin{equation}
    (\widehat{\mathbf{Y}},\, \boldsymbol{\ell}) = B(\mathcal{H},\mathcal{M}),
    \qquad
    \widehat{\mathbf{Y}} \in \mathbb{R}^{A\times K\times T\times 2}.
    \label{eq:proposal_interface}
\end{equation}
For marginal forecasting, $k$ indexes an agent-specific mode and $\ell_{a,k}$ is an agent-level logit. For joint forecasting, $k$ indexes a coherent scene-level future shared by all predicted agents and $\ell_k$ is a scene logit. FutureLift accesses only $\widehat{\mathbf{Y}}$, $\boldsymbol{\ell}$, and the original scene description; it does not consume backbone queries, decoder features, or intermediate activations. We therefore treat $B$ as frozen and stop gradients at the interface in \cref{eq:proposal_interface}. This contract permits the same refiner architecture to operate on structurally different proposal generators after setting-specific training.

FutureLift outputs coordinate and logit residuals,
\begin{equation}
    \widetilde{\mathbf{Y}} = \widehat{\mathbf{Y}} + \Delta\mathbf{Y},
    \qquad
    \widetilde{\boldsymbol{\ell}} = \boldsymbol{\ell} + \Delta\boldsymbol{\ell},
    \label{eq:residual_interface}
\end{equation}
which preserves the backbone's modes while allowing their geometry and ranking to change. The residual parameterization is important for a reusable refiner: the proposal remains the reference solution, and FutureLift learns only the correction supported by the lifted scene context.

\subsection{Proposal-to-Future-Graph Lifting}
\label{subsec:lifting}

\paragraph{Spatially grounded future nodes.}
We partition the horizon into $R$ contiguous intervals $\{\mathcal{I}_r\}_{r=1}^{R}$ and create one future node $v_{a,k,r}$ for every agent, mode, and interval. Hence, lifting produces $AKR$ future nodes without changing the number of trajectory hypotheses. Node $v_{a,k,r}$ is anchored at the proposal endpoint
\begin{align}
    \mathbf{c}_{a,k,r} &= \widehat{\mathbf{y}}_{a,k,t_r}, \\
    \theta_{a,k,r} &= \operatorname{atan2}\!\left(
    [\Delta\widehat{\mathbf{y}}_{a,k,r}]_y,
    [\Delta\widehat{\mathbf{y}}_{a,k,r}]_x
    \right),
    \label{eq:future_anchor}
\end{align}
where $t_r$ is the final step of interval $\mathcal{I}_r$. We estimate headings by finite differences over the endpoint sequence, prepending the last observed position as $\mathbf{c}_{a,k,0}$. Specifically, $\Delta\widehat{\mathbf{y}}_{a,k,r}=(\mathbf{c}_{a,k,r+1}-\mathbf{c}_{a,k,r-1})/2$ for $r<R$, and $\Delta\widehat{\mathbf{y}}_{a,k,R}=\mathbf{c}_{a,k,R}-\mathbf{c}_{a,k,R-1}$ at the final endpoint. Anchoring at the predicted future location is what turns a trajectory array into a spatial hypothesis: neighborhood queries are evaluated where the agent is predicted to be, rather than at its last observation or at a learned, geometry-free query.

Each node also retains the complete proposal segment in its local frame. For $t\in\mathcal{I}_r$, we transform the proposed states as
\begin{equation}
\begin{aligned}
    \mathbf{z}_{a,k,r,t}
    &= \mathbf{R}(\theta_{a,k,r})^{\top}
    \left(\widehat{\mathbf{y}}_{a,k,t}-\mathbf{c}_{a,k,r}\right), \\
    \mathbf{h}^{0}_{a,k,r}
    &= \operatorname{GRU}\!\left(\{\psi(\mathbf{z}_{a,k,r,t})\}_{t\in\mathcal{I}_r}\right),
    \label{eq:segment_embedding}
\end{aligned}
\end{equation}
where $\psi$ is a point-wise Fourier/MLP embedding. Thus, the node position determines where context is retrieved, while its feature preserves the motion pattern within the interval. This separation avoids reducing a segment to its endpoint alone.

\paragraph{Scene nodes and relative geometry.}
The lifted future nodes are combined with a heterogeneous scene graph containing historical agent states, lane elements, and fine-grained map points. FutureLift uses the QCNet scene encoder~\cite{qcnet} to encode this context. The encoder is trained as part of FutureLift from the original scene inputs; FutureLift does not reuse encoder activations or other latent features from the frozen proposal backbone. Following established target-local relative-pose encodings in motion forecasting~\cite{gorela,qcnet,hptr}, all spatial messages use relative geometry. For an edge from source $u$ to target $v$, we embed

\begin{equation}
    \begin{aligned}
        \mathbf{e}_{u \to v}
        &= \phi\Bigl(
        [
        \rho_{uv},\,
        \sin \beta_{uv},\,
        \cos \beta_{uv}, \\
        &\qquad
        \sin \Delta\theta_{uv},\,
        \cos \Delta\theta_{uv},\,
        \Delta t_{uv}
        ]
        \Bigr).
    \end{aligned}
\end{equation}

where $(\rho_{uv},\beta_{uv})$ is the source position in the target's local polar frame, $\Delta\theta_{uv}$ is their heading difference, and $\Delta t_{uv}$ is their temporal offset. Static map relations use $\Delta t_{uv}=0$. Rebuilding these attributes around the proposal anchors makes the lifted graph equivariant to global translation and rotation while retaining proposal-specific geometry.

\paragraph{Typed future relations.}
We define the lifted edge set as
\begin{equation}
    \mathcal{E}_{\mathrm{lift}}
    = \mathcal{E}_{\mathrm{hist}}
    \cup \mathcal{E}_{\mathrm{temp}}
    \cup \mathcal{E}_{\mathrm{map}}
    \cup \mathcal{E}_{\mathrm{social}}
    \cup \mathcal{E}_{\mathrm{mode}}.
    \label{eq:lifted_edges}
\end{equation}
Each relation has a distinct role:
\begin{itemize}
    \item $\mathcal{E}_{\mathrm{hist}}$ connects each future node to its own recent observations and nearby agents, grounding it in motion history and current interactions.
    \item $\mathcal{E}_{\mathrm{temp}}=\{(v_{a,k,r},v_{a,k,s})\mid r\neq s\}$ connects all intervals of a proposal bidirectionally, so an edge from $s>r$ to $r$ carries later evidence to an earlier segment within the same pass.
    \item $\mathcal{E}_{\mathrm{map}}$ connects each future node to nearby lane context and high-resolution map points around $\mathbf{c}_{a,k,r}$, providing route structure and local boundary geometry.
    \item $\mathcal{E}_{\mathrm{social}}$ connects spatially neighboring agents at interval $r$ within joint future $k$. It is disabled for marginal proposals because equal mode indices across agents do not identify corresponding hypotheses and would encode unsupported interactions.
    \item $\mathcal{E}_{\mathrm{mode}}$ connects $v_{a,k,r}$ to $v_{a,k',r}$ for $k'\neq k$, allowing an agent's alternative futures to exchange context without merging their identities.
\end{itemize}
This conditional topology supports both tasks without assigning joint semantics to independent marginal modes.

\subsection{One-Shot Bidirectional Future Refinement}
\label{subsec:refinement}

The lifted graph exposes the complete horizon before coordinate correction. We apply $L$ layers of edge-enhanced, relation-specific multi-head attention~\cite{transformer,hgt,graphtransformer}. Each layer sequentially processes own-history, bidirectional future-temporal, map, and social relations, skipping disabled relations. Each concrete relation has its own attention parameters, gate, normalization layers, and FFN, and immediately supplies its updated future-node features to the next operator.

For an operator of relation $q$, let $\bar{\mathbf{h}}_v$, $\bar{\mathbf{h}}_u$, and $\bar{\mathbf{e}}_{u\to v}$ denote its layer-normalized current target, source, and edge features. Queries come from future targets; keys and values come from the relation's source nodes and relative geometry. One head of dimension $d_h$ computes
\begin{align}
    s_{uv}^{q}&=\frac{\bigl(\mathbf{W}_{Q}^{q}\bar{\mathbf{h}}_{v}\bigr)^{\top}\bigl(\mathbf{W}_{K}^{q}\bar{\mathbf{h}}_{u}
    +\mathbf{W}_{K_E}^{q}\bar{\mathbf{e}}_{u\to v}\bigr)}{\sqrt{d_h}}, \\
    \alpha_{uv}^{q}&=\operatorname{softmax}_{u\in\mathcal{N}_{q}(v)}\left(s_{uv}^{q}\right), \\
    \mathbf{m}^{q}_{v} &= \sum_{u\in\mathcal{N}_{q}(v)} \alpha^{q}_{uv}
    \left(\mathbf{W}^{q}_{V}\bar{\mathbf{h}}_{u}+\mathbf{W}^{q}_{V_E}\bar{\mathbf{e}}_{u\rightarrow v}\right),
    \label{eq:relation_attention}
\end{align}
where head and layer indices are suppressed and projection biases are implicit. The heads are concatenated, mixed with a projected target feature by a channel-wise gate, and output-projected. Each operator completes its attention and FFN residual updates before the next relation reads the resulting state; \cref*{appendix:sec:attention_operator} specifies the gate and normalization equations. Cross-mode attention is applied only after the final layer's social operators.

The phrase \emph{one-shot} refers to a single refinement pass on a fixed proposal graph. Relation types are processed sequentially, while all temporal intervals remain available throughout. Since temporal attention precedes map and social attention within a layer, evidence retrieved at a later anchor can reach an earlier interval through temporal attention in the following layer. Coordinates for all intervals are decoded after message passing, with no causal segment sweep, intermediate coordinate update, or graph reconstruction. Additional outer refinement passes are evaluated in \cref*{appendix:tab:refinement_passes}.

After message passing, a type-specific MLP decodes a local coordinate correction and raw uncertainty parameters for every interval. We rotate the coordinate corrections back to the trajectory frame and concatenate them in temporal order to form $\Delta\mathbf{Y}$ in \cref{eq:residual_interface}; positive Laplace scales are obtained as $\widetilde{\mathbf{b}}=\operatorname{ELU}(\mathbf{b}_{\mathrm{raw}})+1+\epsilon$, with $\epsilon=0.1$. For marginal prediction, the updated states of one agent--mode pair are concatenated and decoded into $\Delta\ell_{a,k}$. For joint prediction, a final future-to-scene attention operator pools the agents and temporal intervals belonging to scene hypothesis $k$, without edge features, before decoding the shared scene-logit residual $\Delta\ell_k$. Coordinate, uncertainty, and confidence refinement therefore use the same proposal-conditioned future representation.

\subsection{Training Objective}
\label{subsec:training}

We train FutureLift with winner-takes-all trajectory regression and likelihood-based mode scoring while keeping proposal generation fixed. Let $\widetilde{\mathbf{Y}}^{k}$ denote a refined mode and $\mathbf{Y}$ the ground truth. In marginal forecasting, the winning mode is selected independently for each agent,
\begin{equation}
    k_a^{*}=\arg\min_{k}
    \frac{1}{|\mathcal{T}_a|}\sum_{t\in\mathcal{T}_a}
    \left\|\widetilde{\mathbf{y}}_{a,k,t}-\mathbf{y}_{a,t}\right\|_2,
    \label{eq:marginal_assignment}
\end{equation}
where $\mathcal{T}_a$ contains valid future steps. In joint forecasting, we select one coherent future scene by aggregating the displacement error over all valid agents,
\begin{equation}
    k^{*}=\arg\min_{k}
    \frac{1}{\sum_a |\mathcal{T}_a|}
    \sum_{a}\sum_{t\in\mathcal{T}_a}
    \left\|\widetilde{\mathbf{y}}_{a,k,t}-\mathbf{y}_{a,t}\right\|_2.
    \label{eq:joint_assignment}
\end{equation}
For trajectory regression, we minimize the coordinate-wise Laplace negative log-likelihood of the selected mode, $|y-\widetilde{y}|/\widetilde{b}+\log(2\widetilde{b})$, summed over the $x$ and $y$ coordinates and averaged over valid agents and time steps. For score refinement, we minimize the negative log-likelihood of the endpoint Laplace mixture over all modes. This loss uses the refined softmax probabilities as mixture weights and stops gradients through the endpoint Laplace terms, so it updates only the refined logits. In marginal forecasting the score loss is computed per agent; in joint forecasting the endpoint likelihood is first averaged over valid agents in each scene. Both losses have unit weight. We also stop gradients through winner selection, proposal coordinates, and proposal logits. Consequently, training cannot modify the proposal mechanism or rely on backbone-specific latent features; it learns only how to lift and correct the supplied hypotheses.

\paragraph{Temporal resolution.}
We use $R=3$ temporal nodes for the six-second Argoverse~2 horizon and $R=4$ for the eight-second Waymo horizon; all remaining implementation and training details are provided in \cref*{appendix:sec:experimentsetup}.

\section{Evaluation}
\label{sec:evaluation}

We organize the evaluation around six questions. First, is FutureLift's proposal-only interface compatible with structurally different forecasting backbones? Second, how does it compare with native and specialized refiners when the proposal backbone is controlled? Third, do the gains hold in marginal and joint prediction and on both Argoverse~2 (AV2) and the Waymo Open Motion Dataset (WOMD) after setting-specific training? Fourth, which relations in the lifted future graph are responsible for the improvement? Fifth, how does the temporal resolution of the lifted graph trade forecasting accuracy against refinement latency? Sixth, how large are the corrections, do substantial corrections improve the same trajectory hypothesis, and how does refinement affect off-road and collision rates? We conclude with representative failure cases; qualitative corrections are provided in \cref*{appendix:subsec:qualitative}. Unless otherwise stated, validation results for refiners trained in our study are means over the same three random seeds; metric-wise standard deviations do not exceed $3\times10^{-3}$. Published comparison rows are marked separately. The seed values and full training protocol are provided in \cref*{appendix:sec:experimentsetup}.

\subsection{AV2 Marginal Prediction}
\label{subsec:av2_marginal}

We first isolate the effect of the refinement module on the AV2 marginal task. We evaluate proposal sets from QCNet~\cite{qcnet}, DenseTNT~\cite{densetnt}, and DONUT~\cite{donut}, comparing the proposal-stage output with each available native or specialized refiner in \cref{tab:av2_marginal_refinement}. FutureLift and the unrefined or native baselines within each block use the same frozen proposal coordinates and logits. Following its official implementation, SmartRefine additionally receives the corresponding backbone trajectory features. Its QCNet row uses the authors' matched-checkpoint export rather than our bit-identical proposal file and is marked with $\dagger$; the DenseTNT comparison uses an identical proposal stage. FutureLift replaces rather than follows the native refiner and receives no backbone latent features.

\begin{table}[h]
    \centering
    \caption{Backbone-interface compatibility on the AV2 marginal validation split. Displacement errors are in meters; runtime is reported in \cref*{appendix:tab:detailed_efficiency}. $^\dagger$SmartRefine result using the authors' matched-checkpoint export and backbone trajectory features.}
    \label{tab:av2_marginal_refinement}
    \scriptsize
    \setlength{\tabcolsep}{3pt}
    \begin{tabular}{@{}llcccc@{}}
        \toprule
        Backbone & Refiner & b-minFDE$_6$ $\downarrow$ & minFDE$_6$ $\downarrow$ & minADE$_6$ $\downarrow$ & MR$_6$ $\downarrow$ \\
        \midrule
        \multirow{4}{*}{QCNet~\cite{qcnet}} & None & 1.912 & 1.304 & 0.729 & 0.164 \\
        & Native & 1.856 & 1.253 & 0.720 & 0.157 \\
        & SmartRefine$^\dagger$ & 1.852 & 1.258 & 0.718 & 0.157 \\
        & \textbf{FutureLift} & \textbf{1.772} & \textbf{1.124} & \textbf{0.593} & \textbf{0.131} \\
        \midrule
        \multirow{3}{*}{DenseTNT~\cite{densetnt}} & None & 2.231 & 1.624 & 0.964 & 0.233 \\
        & SmartRefine & 2.109 & 1.553 & 0.834 & 0.221 \\
        & \textbf{FutureLift} & \textbf{1.996} & \textbf{1.382} & \textbf{0.718} & \textbf{0.161} \\
        \midrule
        \multirow{3}{*}{DONUT~\cite{donut}} & None & 1.838 & 1.193 & 0.728 & 0.146 \\
        & Native & 1.807 & 1.176 & 0.722 & 0.144 \\
        & \textbf{FutureLift} & \textbf{1.742} & \textbf{1.097} & \textbf{0.582} & \textbf{0.124} \\
        \bottomrule
    \end{tabular}
\end{table}

FutureLift improves every metric for all three proposal backbones. Relative to its shared unrefined proposals, it reduces minFDE by 13.8\%, 14.9\%, and 8.0\% on QCNet, DenseTNT, and DONUT, respectively; the corresponding minADE reductions are 18.7\%, 25.5\%, and 20.1\%. Relative to the matched QCNet and identical-stage DenseTNT baselines, SmartRefine reduces minFDE by 3.5\% and 4.4\% and minADE by 1.5\% and 13.5\%, respectively. FutureLift's metrics are respectively 10.7\% and 11.0\% lower in minFDE and 17.4\% and 13.9\% lower in minADE than SmartRefine's. The identical-input FutureLift gains across QCNet, DenseTNT, and DONUT demonstrate compatibility of its proposal-only interface; the specialized-refiner comparisons additionally control the backbone setting but, where marked, not the exact export or input contract.

\subsection{AV2 Joint Prediction}
\label{subsec:av2_joint}

In \cref{tab:av2_joint_refinement} we test whether the same proposal-only FutureLift interface is compatible with joint forecasting. FutureLift is attached to frozen FJMP~\cite{fjmp}, Forecast-MAE~\cite{cheng2023forecast}, and DONUT proposal models and trained separately for each backbone. The SRefiner~\cite{srefiner} rows reproduce the authors' published results for matched FJMP and Forecast-MAE backbones; these use the authors' proposal exports rather than our identical proposal files. Moreover, following its released AV2 implementation, SRefiner receives a precomputed 134-dimensional per-mode embedding in addition to trajectory coordinates and map context. We mark these protocol differences with $\dagger$ and interpret SRefiner as a matched-backbone specialized-refiner comparison rather than a proposal-only interface comparison. The DONUT block uses a stronger proposal distribution, so raw metrics should be compared within rather than across backbone blocks. 

\begin{table}[h]
    \centering
    \caption{Backbone-interface compatibility on the AV2 joint validation split. Displacement errors are in meters; runtime is reported in \cref*{appendix:tab:detailed_efficiency}. $^\dagger$Published result using the SRefiner authors' matched-backbone proposal export and additional per-mode embeddings.}
    \label{tab:av2_joint_refinement}
    \scriptsize
    \setlength{\tabcolsep}{3pt}
    \begin{tabular}{@{}llcccc@{}}
        \toprule
        Backbone & Refiner & \shortstack{avgBrier-\\minFDE$_6$ $\downarrow$} & \shortstack{avgMin\\FDE$_6$ $\downarrow$} & \shortstack{avgMin\\ADE$_6$ $\downarrow$} & \shortstack{Actor\\MR$_6$ $\downarrow$} \\
        \midrule
        \multirow{3}{*}{FJMP~\cite{fjmp}} & None & 2.478 & 1.921 & 0.812 & 0.237 \\
        & SRefiner$^\dagger$ & 2.354 & 1.736 & 0.747 & 0.221 \\
        & \textbf{FutureLift} & \textbf{2.114} & \textbf{1.512} & \textbf{0.699} & \textbf{0.187} \\
        \midrule
        \multirow{3}{*}{\shortstack{Forecast-\\MAE}~\cite{cheng2023forecast}} & None & 2.412 & 1.642 & 0.713 & 0.195 \\
        & SRefiner$^\dagger$ & 2.251 & 1.477 & 0.658 & 0.183 \\
        & \textbf{FutureLift} & \textbf{2.012} & \textbf{1.391} & \textbf{0.642} & \textbf{0.169} \\
        \midrule
        \multirow{3}{*}{DONUT~\cite{donut}} & None & 1.944 & 1.262 & 0.584 & 0.162 \\
        & FutureLift w/o social & 1.831 & 1.157 & 0.561 & 0.148 \\
        & \textbf{FutureLift} & \textbf{1.816} & \textbf{1.142} & \textbf{0.552} & \textbf{0.146} \\
        \bottomrule
    \end{tabular}
\end{table}

\begin{table*}[t]
    \centering
    \caption{WOMD evaluation. (a) Official test metrics averaged over agent types and the 3, 5, and 8-second horizons, except the all-object minFDE@8s. FutureLift uses one randomly selected trained seed; bold denotes the best standard-input result. Protocol-different systems are shown only for context. (b) Horizon-resolved validation mean and standard deviation over three seeds; each model is evaluated at all horizons without retraining.}
    \label{tab:womd_generalization}
    \scriptsize
    \setlength{\tabcolsep}{3pt}
    \textbf{(a) Official WOMD test results}\\[2pt]
    \begin{tabular}{@{}lccccccc@{}}
        \toprule
        Method & Soft mAP $\uparrow$ & mAP $\uparrow$ & minADE $\downarrow$ (m) & minFDE $\downarrow$ (m) & minFDE@8s $\downarrow$ (m) & MR $\downarrow$ & OR $\downarrow$ \\
        \midrule
        \multicolumn{8}{@{}l}{\emph{Published methods with standard WOMD inputs}} \\
        \addlinespace[1pt]
        MTR++~\cite{mtr++} & 0.4410 & 0.4329 & 0.5906 & 1.1939 & -- & 0.1298 & 0.1281 \\
        ModeSeq~\cite{modeseq} & 0.4487 & 0.4450 & 0.5304 & 1.0836 & 2.153 & 0.1244 & -- \\
        NDPNet~\cite{ndpnet} & 0.4504 & 0.4441 & 0.5164 & 1.0321 & 1.7513 & 0.1238 & 0.1291 \\
        IMPACT~\cite{impact} & 0.4721 & 0.4609 & 0.5641 & 1.1540 & 1.9563 & \textbf{0.1143} & \textbf{0.1255} \\
        DONUT~\cite{donut} & 0.2582 & 0.2321 & 0.7676 & 1.0539 & 1.7942 & 0.1229 & 0.1733 \\
        \textbf{FutureLift} (DONUT) & \textbf{0.4874} & \textbf{0.4780} & \textbf{0.5049} & \textbf{0.9942} & \textbf{1.6815} & 0.1146 & 0.1283 \\
        \midrule
        \multicolumn{8}{@{}l}{\emph{Protocol-different leaderboard context}} \\
        \addlinespace[1pt]
        Modeseq* \cite{modeseq} & 0.4737 & 0.4665 & 0.5680 & 1.1766 & 2.0072 & 0.1204 & 0.1275 \\
        IMPACT* \cite{impact} & 0.4801 & 0.4598 & 0.5563 & 1.1295 & 1.9131 & 0.1087 & 0.1258 \\
        MTR v3$^{*\dagger}$~\cite{mtrv3} & 0.4967 & 0.4859 & 0.5554 & 1.1062 & 1.8500 & 0.1098 & 0.1279 \\
        DONUT-NLL (window policy)$^{\ddagger}$~\cite{metricagnostic} & 0.5018 & 0.4987 & -- & 1.3280 & 2.2270 & 0.0900 & 0.1127 \\
        \bottomrule
    \end{tabular}

    \vspace{1pt}
    {\raggedright\scriptsize $^*$Ensemble. $^\dagger$Additional LiDAR input. $^\ddagger$Metric-specific output policy; the anomalous leaderboard minADE is excluded because the cited comparison does not report a valid value.\par}

    \vspace{5pt}
    \textbf{(b) WOMD validation by evaluation horizon}\\[2pt]
    \begin{tabular}{@{}lcccccc@{}}
        \toprule
        \multicolumn{7}{c}{\tiny SD after $\pm$ is reported in units of $10^{-3}$ where shown.} \\
        \addlinespace[1pt]
        & \multicolumn{2}{c}{3 s} & \multicolumn{2}{c}{5 s} & \multicolumn{2}{c}{8 s} \\
        \cmidrule(lr){2-3}\cmidrule(lr){4-5}\cmidrule(l){6-7}
        Method & minADE $\downarrow$ & minFDE $\downarrow$ & minADE $\downarrow$ & minFDE $\downarrow$ & minADE $\downarrow$ & minFDE $\downarrow$ \\
        \midrule
        DONUT (native) & 0.3588 & 0.4421 & 0.7571 & 0.9036 & 1.1545 & 1.7853 \\
        DONUT + FutureLift (causal) & \textbf{0.2373{\scriptsize$\pm\mathrm{0.2}$}} & \textbf{0.4210{\scriptsize$\pm\mathrm{0.2}$}} & 0.4538{\scriptsize$\pm\mathrm{0.1}$} & 0.8737{\scriptsize$\pm\mathrm{0.6}$} & 0.8160{\scriptsize$\pm\mathrm{0.3}$} & 1.7443{\scriptsize$\pm\mathrm{1.2}$} \\
        DONUT + FutureLift (bidirectional) & 0.2401{\scriptsize$\pm\mathrm{0.4}$} & 0.4217{\scriptsize$\pm\mathrm{0.4}$} & \textbf{0.4515{\scriptsize$\pm\mathrm{0.2}$}} & \textbf{0.8498{\scriptsize$\pm\mathrm{0.5}$}} & \textbf{0.7981{\scriptsize$\pm\mathrm{0.4}$}} & \textbf{1.6666{\scriptsize$\pm\mathrm{1.4}$}} \\
        \bottomrule
    \end{tabular}
\end{table*}
Relative to the shared unrefined proposals, FutureLift reduces avgMinFDE by 21.3\%, 15.3\%, and 9.5\% on FJMP, Forecast-MAE, and DONUT, respectively; the corresponding ActorMR reductions are 21.1\%, 13.3\%, and 9.9\%. Relative to the matched FJMP and Forecast-MAE baselines reported by its authors, SRefiner reduces avgMinFDE by 9.6\% and 10.0\% and ActorMR by 6.8\% and 6.2\%, respectively. FutureLift's resulting metrics are 12.9\% and 5.8\% lower in avgMinFDE and 15.4\% and 7.7\% lower in ActorMR than the corresponding published SRefiner metrics. Because SRefiner uses a different proposal export and additional per-mode embeddings, these differences compare matched backbone settings but do not isolate the refiner under an identical proposal-only input contract. On DONUT, adding social future relations to FutureLift further reduces avgMinFDE by 1.3\% and ActorMR by 1.4\%.
Because the AV2 leaderboard is closed to new submissions, we report locally evaluated marginal and joint test-set comparisons in \cref*{appendix:tab:av2_marginal_context,appendix:tab:av2_joint_context}.

\subsection{WOMD Evaluation and Horizon Dependence}
\label{subsec:womd}

WOMD evaluates FutureLift after setting-specific training on a second dataset with an eight-second horizon. Panel~(a) of \cref{tab:womd_generalization} reports official test results; FutureLift uses one model selected randomly from three trained seeds, and protocol-different systems are separated as leaderboard context. Panel~(b) compares causal and bidirectional variants on the validation split, reporting mean and standard deviation over three seeds. For each variant and seed, the same model is evaluated at 3, 5, and 8 seconds without horizon-specific retraining.

FutureLift achieves the lowest valid horizon-averaged minADE and minFDE among the listed standard-input methods, with competitive Soft mAP. Its single model also outperforms all listed ensembles on every reported displacement metric. At 8~s, its 1.6815~m minFDE improves over the next-best standard-input result, NDPNet's 1.7513~m, by 4.0\%. DONUT-NLL uses metric-specific output selection~\cite{metricagnostic} and therefore serves only as leaderboard context. Relative to the causal graph, bidirectional reasoning increases minADE/minFDE by 1.2\%/0.2\% at 3~s, but reduces them by 0.5\%/2.7\% at 5~s and 2.2\%/4.5\% at 8~s. This growing advantage supports backward temporal information flow over longer horizons.

\subsection{What Makes Future-Graph Lifting Effective?}
\label{subsec:core_ablation}

We build FutureLift cumulatively on one frozen DONUT checkpoint on the AV2 joint validation split. Starting from the unrefined proposals, we add history, map, and cross-mode context. We then compare one to three forward-only causal sweeps with one bidirectional sweep before introducing the joint-specific social topology. Here, ``x$J$'' denotes $J$ repeated causal sweeps, each reconstructing the graph from the latest coordinates. This design tests whether direct backward information flow adds value beyond repeated forward-only refinement.

\begin{table}[t]
    \centering
    \caption{FutureLift component build-up on the AV2 joint validation split using one frozen DONUT checkpoint. Context relations are added cumulatively; the temporal rows compare $J$ repeated causal sweeps (x$J$) with one bidirectional sweep.}
    \label{tab:core_ablation}
    \scriptsize
    \setlength{\tabcolsep}{3.1pt}
    \begin{tabular}{@{}lcccc@{}}
        \toprule
        \multicolumn{5}{c}{\tiny SD after $\pm$ is reported in units of $10^{-3}$ where shown.} \\
        \addlinespace[1pt]
        Configuration & \shortstack{avgBrier-\\minFDE$_6$ $\downarrow$} & \shortstack{avgMin\\FDE$_6$ $\downarrow$} & \shortstack{avgMin\\ADE$_6$ $\downarrow$} & \shortstack{Actor\\MR$_6$ $\downarrow$} \\
        \midrule
        No refinement & 1.944 & 1.262 & 0.584 & 0.162 \\
        + History grounding & 1.924 & 1.253 & 0.581 & 0.158 \\
        + Map grounding & 1.886 & 1.216 & 0.572 & 0.154 \\
        + Cross-mode grounding & 1.862 & 1.204 & 0.570 & 0.153 \\
        + Causal temp. reasoning x1 & 1.857 & 1.199{\scriptsize$\pm\mathrm{1.5}$} & 0.569 & 0.152{\scriptsize$\pm\mathrm{0.1}$} \\
        + Causal temp. reasoning x2 & 1.853 & 1.191{\scriptsize$\pm\mathrm{1.2}$} & 0.568 & 0.152{\scriptsize$\pm\mathrm{0.1}$} \\
        + Causal temp. reasoning x3 & 1.851 & 1.191{\scriptsize$\pm\mathrm{1.4}$} & 0.567 & 0.151{\scriptsize$\pm\mathrm{0.1}$} \\
        + Bidirectional temp. reasoning & 1.821 & 1.157{\scriptsize$\pm\mathrm{1.1}$} & 0.561 & 0.148{\scriptsize$\pm\mathrm{0.1}$} \\
        + Social future relations (full) & \textbf{1.816} & \textbf{1.142{\scriptsize$\pm\mathrm{1.3}$}} & \textbf{0.552} & \textbf{0.146{\scriptsize$\pm\mathrm{0.1}$}} \\
        \bottomrule
    \end{tabular}
\end{table}

Repeated causal computation saturates after two sweeps, see \cref{tab:core_ablation}: a third sweep leaves avgMinFDE at 1.191~m. In contrast, one bidirectional sweep reduces avgMinFDE to 1.157~m and ActorMR from 0.151 to 0.148, showing that direct backward temporal edges are more effective than up to three forward-only sweeps in this controlled configuration. Adding social future relations then yields a further 0.015~m avgMinFDE reduction and completes the joint model.

\subsection{Temporal Resolution}

The number of lifted intervals controls the trade-off between temporal granularity and graph cost. At $R=1$, the complete six-second proposal is compressed into one trajectory-level token, represented by a single spatially anchored future node. This baseline retains separate relation-specific attention to map context, agent histories, alternative modes, and other agents' trajectories, but has no inter-segment temporal edges. We vary $R$ while keeping the DONUT checkpoint, FutureLift capacity, and AV2 joint training protocol fixed.

\begin{table}[h]
    \centering
    \caption{Temporal-resolution sensitivity on the AV2 joint validation split. $R=1$ is the whole-trajectory-token baseline; the default is $R=3$. Latency excludes the backbone and is measured on an RTX 6000 Ada GPU.}
    \label{tab:temporal_resolution}
    \scriptsize
    \setlength{\tabcolsep}{2.8pt}
    \begin{tabular}{@{}lccccc@{}}
        \toprule
        \multicolumn{6}{c}{\tiny SD after $\pm$ is reported in units of $10^{-3}$ where shown.} \\
        \addlinespace[1pt]
        Resolution & \shortstack{avgBrier-\\minFDE$_6$ $\downarrow$} & \shortstack{avgMin\\FDE$_6$ $\downarrow$} & \shortstack{avgMin\\ADE$_6$ $\downarrow$} & \shortstack{Actor\\MR$_6$ $\downarrow$} & \shortstack{Refinement\\latency (ms)} \\
        \midrule
        $R=1$ (6 s / node) & 1.884 & 1.221 & 0.599 & 0.161 & \textbf{29} \\
        $R=2$ (3 s / node) & 1.827 & 1.162 & 0.574 & 0.153 & 35 \\
        $R=3$ (2 s / node) & 1.816 & 1.142{\scriptsize$\pm1.0$} & 0.552{\scriptsize$\pm0.8$} & 0.146 & 42 \\
        $R=6$ (1 s / node) & \textbf{1.810} & \textbf{1.139{\scriptsize$\pm0.9$}} & \textbf{0.551{\scriptsize$\pm0.6$}} & \textbf{0.144} & 52 \\
        \bottomrule
    \end{tabular}
\end{table}

Compared with the whole-trajectory-token baseline ($R=1$), segment-level lifting at $R=3$ reduces avgMinFDE from 1.221 to 1.142~m and avgMinADE from 0.599 to 0.552~m. Increasing the resolution from $R=3$ to $R=6$ reduces avgMinFDE by only 0.003~m and avgMinADE by 0.001~m while increasing refinement latency from 42 to 52~ms. We retain $R=3$ as the default, capturing most of the accuracy gain at a lower refinement cost.

\subsection{Correction Magnitude}
\label{subsec:correction_magnitude}

\begin{figure}[t]
    \centering
    \includegraphics[width=\columnwidth]{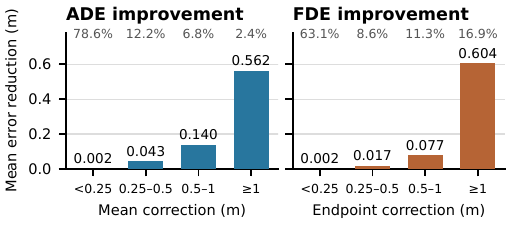}
    \caption{Mean ADE/FDE reduction by Euclidean correction magnitude over valid future steps. We retain the original highest-scoring mode (ties broken by index) for 97,799 agents. Percentages indicate bin populations; positive gains mean lower error.}
    \label{fig:correction_gain}
\end{figure}

On the AV2 joint validation split with DONUT proposals, larger corrections are associated with greater same-mode error reductions (\cref{fig:correction_gain}). Although most corrections are small, 16.9\% of original-mode endpoints move at least 1~m; FDE improves for 62.1\% of these trajectories.
Across all modes, off-road trajectory rate decreases from 5.749\% to 3.953\%, whereas pair collision rate decreases slightly from 3.846\% to 3.353\% (protocol in \cref*{appendix:subsec:geometric_diagnostics}). Improved displacement accuracy therefore does not guarantee fewer collisions.

\subsection{Limitations and Failure Modes}
\label{subsec:limitations}

\begin{figure}[h]
    \centering
    \begin{subfigure}{0.49\linewidth}
        \centering
        \includegraphics[height=4.5cm]{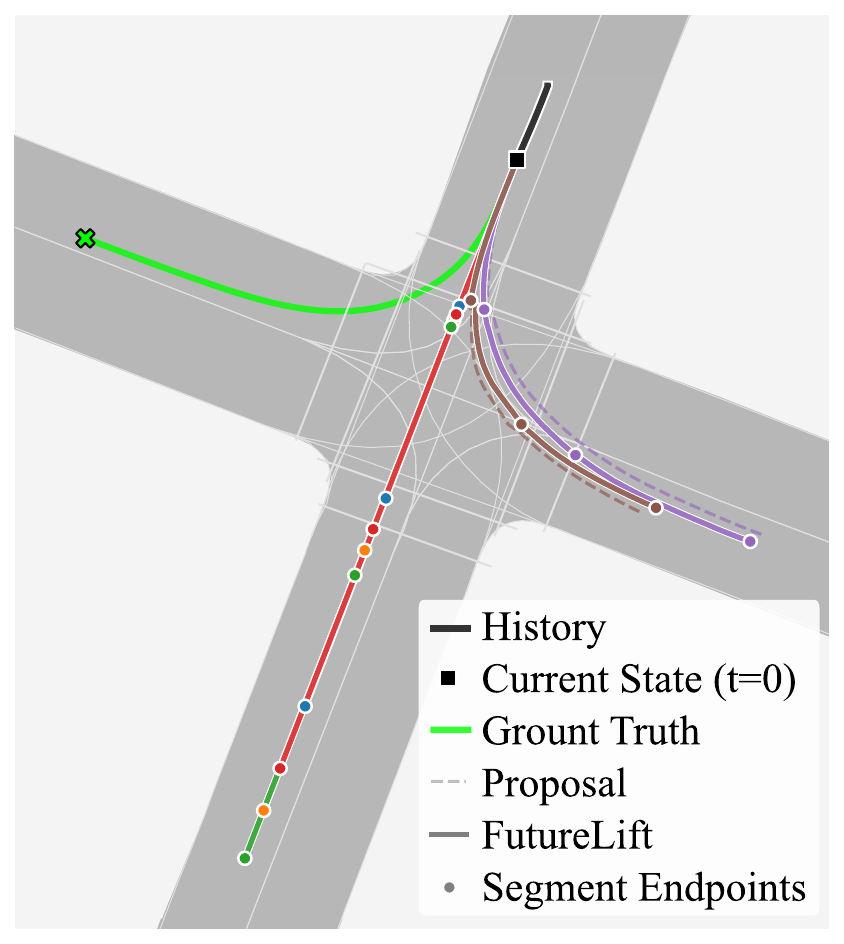}
        \caption{Mode Collapse}
        \label{fig:limit2}
    \end{subfigure}
    \begin{subfigure}{0.49\linewidth}
        \centering
        \includegraphics[height=4.5cm]{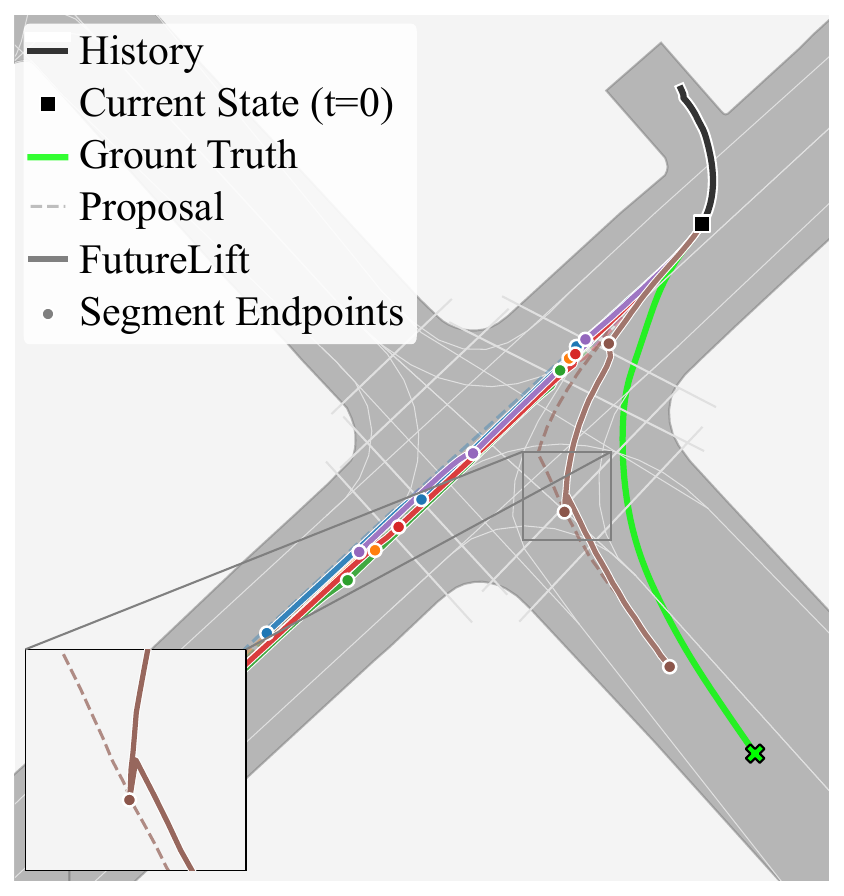}
        \caption{Kinematic Feasibility}
        \label{fig:limit1}
    \end{subfigure}
    \caption{\textbf{Representative failure cases.} \textbf{Left:} The frozen proposal set omits the correct maneuver; post-hoc refinement cannot recover a mode absent from its inputs. \textbf{Right:} FutureLift produces excessive curvature because its residual decoder does not explicitly enforce vehicle dynamics.}
    \label{fig:limits}
\end{figure}

FutureLift inherits important limitations from post-hoc proposal refinement. First, it preserves the number of input modes and predicts residual corrections to their coordinates and scores. It therefore cannot reliably recover a maneuver that is absent from the frozen proposal set, as illustrated in \cref {fig:limit2}. Second, the coordinate decoder does not explicitly enforce vehicle dynamics, so aggressive corrections can occasionally produce excessive curvature (\cref{fig:limit1}). Finally, our evaluation uses open-loop forecasting metrics and does not establish that the observed geometric improvements translate directly to closed-loop planning or safety. These limitations motivate dynamics-aware residual constraints, proposal-diversity mechanisms, and planner-level evaluation.

\section{Conclusion}
\label{sec:conclusion}

We introduced FutureLift, a backbone-agnostic refiner that lifts multimodal trajectory proposals into spatially grounded future graphs. Bidirectional temporal relations let later context revise earlier segments in one graph pass. Across Argoverse~2 and WOMD, separately trained refiners consistently improve frozen proposals from structurally different backbones in marginal and joint forecasting. FutureLift currently ranks first in both minADE (0.5049~m) and minFDE (0.9942~m) on the WOMD motion-prediction leaderboard. Ablations show that segment-level lifting outperforms whole-trajectory tokens and that bidirectional reasoning outperforms repeated causal sweeps, with a growing advantage over causal reasoning at longer WOMD horizons. Extending FutureLift to map-light settings and testing whether its forecasting gains improve closed-loop planning remain future work.

\section*{Acknowledgements}
This paper was created in the ``Country 2 City- Bridge'' project of the ``German Center for Future Mobility'', which is funded by the German Federal Ministry of Transport. Responsibility for the information and views set out in this publication lies entirely with the authors.

\clearpage
{
    \small
    \bibliographystyle{ieeenat_fullname}
    \bibliography{main}
}
\clearpage
\maketitlesupplementary
\appendix
\begin{table*}[t]
    \centering
    \caption{Additional AV2 marginal test-set context. Asterisks denote ensembles. FutureLift is evaluated locally with the released test annotations and official evaluator because the leaderboard is closed to new submissions. These results complement, but are distinct from, the controlled validation comparisons in \cref*{tab:av2_marginal_refinement}.}
    \label{appendix:tab:av2_marginal_context}
    \scriptsize
    \setlength{\tabcolsep}{4pt}
    \begin{tabular}{@{}lcccccc@{}}
        \toprule
        Method & b-minFDE$_6$ $\downarrow$ (m) & minFDE$_6$ $\downarrow$ (m) & minADE$_6$ $\downarrow$ (m) & MR$_6$ $\downarrow$ & minFDE$_1$ $\downarrow$ (m) & minADE$_1$ $\downarrow$ (m) \\
        \midrule
        QCNet~\cite{qcnet} & 1.91 & 1.29 & 0.65 & 0.16 & 4.30 & 1.69 \\
        SmartRefine~\cite{smartrefine} & 1.86 & 1.23 & 0.63 & 0.15 & 4.17 & 1.65 \\
        DeMo~\cite{demo} & 1.84 & 1.17 & 0.61 & 0.13 & 3.74 & 1.49 \\
        Polaris~\cite{polaris} & 1.80 & 1.15 & 0.62 & 0.13 & 3.78 & 1.53 \\
        DONUT~\cite{donut} & 1.79 & 1.16 & 0.63 & 0.14 & 4.06 & 1.66 \\
        NDPNet~\cite{ndpnet} & 1.83 & 1.17 & 0.61 & 0.14 & 4.08 & 1.62 \\
        DeMo++~\cite{demo++} & 1.74 & 1.12 & 0.61 & \textbf{0.12} & \textbf{3.70} & 1.50 \\
        \midrule
        \textbf{FutureLift} (DONUT) & 1.74 & \textbf{1.09} & \textbf{0.58} & \textbf{0.12} & 3.90 & 1.56 \\
        \midrule
        QCNet* \cite{qcnet} & 1.78 & 1.19 & 0.62 & 0.14 & 3.96 & 1.56 \\
        SEPT* \cite{sept} & 1.74 & 1.15 & 0.61 & 0.14 & 3.70 & 1.48 \\
        DeMo* \cite{demo} & 1.73 & 1.11 & 0.60 & 0.12 & 3.70 & 1.49 \\
        Polaris* \cite{polaris} & 1.71 & 1.11 & 0.61 & 0.12 & 3.78 & 1.53 \\
        FutureNet-LOF* \cite{futurenet} & 1.63 & 1.07 & 0.58 & 0.12 & 3.63 & 1.46 \\
        NDPNet*~\cite{ndpnet} & \textbf{1.71} & \textbf{1.09} & \textbf{0.58} & 0.13 & 3.71 & \textbf{1.47} \\
        \bottomrule
    \end{tabular}
\end{table*}

\begin{table*}[t]
    \centering
    \caption{Additional AV2 joint test-set context. FutureLift is evaluated locally with the released test annotations and official evaluator because the leaderboard is closed to new submissions. These results complement, but are distinct from, the controlled validation comparisons in \cref*{tab:av2_joint_refinement}.}
    \label{appendix:tab:av2_joint_context}
    \scriptsize
    \setlength{\tabcolsep}{4pt}
    \begin{tabular}{@{}lcccccc@{}}
        \toprule
        Method & avgBrier-minFDE$_6$ $\downarrow$ (m) & avgMinFDE$_6$ $\downarrow$ (m) & avgMinADE$_6$ $\downarrow$ (m) & ActorMR$_6$ $\downarrow$ & avgMinFDE$_1$ $\downarrow$ (m) & avgMinADE$_1$ $\downarrow$ (m) \\
        \midrule
        FJMP~\cite{fjmp} & 2.59 & 1.89 & 0.81 & 0.23 & 4.00 & 1.52 \\
        Forecast-MAE~\cite{cheng2023forecast} & 2.24 & 1.55 & 0.69 & 0.19 & 3.33 & 1.30 \\
        RealMotion~\cite{realmotion} & 2.01 & 1.32 & 0.62 & 0.18 & 2.87 & 1.14 \\
        DeMo~\cite{demo} & 1.93 & 1.24 & 0.58 & 0.16 & 2.78 & 1.12 \\
        Polaris~\cite{polaris} & 1.86 & 1.18 & 0.56 & \textbf{0.15} & 2.67 & \textbf{1.07} \\
        \midrule
        \textbf{FutureLift} (DONUT) & \textbf{1.82} & \textbf{1.14} & \textbf{0.55} & \textbf{0.15} & \textbf{2.63} & \textbf{1.07} \\
        \bottomrule
    \end{tabular}
\end{table*}

\section{Experimental Setup}
\label{appendix:sec:experimentsetup}

\subsection{Datasets and Tasks}

Argoverse~2 (AV2) contains 199,908 training, 24,988 validation, and 24,984 test scenarios~\cite{argoverse2}. Each scenario provides five seconds of history and requires six seconds of prediction at 10 Hz. Models predict the future positions of the scored actors. In marginal forecasting, each actor receives an independently ranked set of trajectories; in joint forecasting, each mode is a coherent scene-level world containing one trajectory per scored actor. All AV2 experiments use $K=6$ modes, while the $K=1$ appendix metrics evaluate the highest-confidence mode.

The Waymo Open Motion Dataset (WOMD) contains 486,995 training, 44,097 validation, and 44,920 test scenarios~\cite{womd}. It provides one second of history and an eight-second prediction horizon at 10 Hz, and designates up to eight agents per scenario for evaluation. We use the marginal prediction setting with $K=6$ trajectories per designated agent. The official metrics are computed separately for vehicles, pedestrians, and cyclists at evaluation cutoffs of 3, 5, and 8 seconds and then averaged. We additionally retain the per-cutoff validation metrics to test whether refinement gains change with horizon length.

\subsection{Metrics}

Let $\mathbf{y}_{a,t}$ and $\hat{\mathbf{y}}^{k}_{a,t}$ denote the ground-truth and mode-$k$ positions of actor $a$ at future step $t$. Average displacement error (ADE) averages $\lVert\hat{\mathbf{y}}^{k}_{a,t}-\mathbf{y}_{a,t}\rVert_2$ over the prediction horizon, whereas final displacement error (FDE) evaluates it at the final step. For AV2 marginal prediction, the mode with the smallest FDE is selected; minFDE and minADE report its FDE and ADE, respectively. MR is the fraction of actors for which every mode has an FDE greater than 2~m. Brier-minFDE additionally accounts for confidence as $\mathrm{minFDE}+(1-s_{k^*})^2$, where $s_{k^*}$ is the normalized confidence of the selected mode. Unless otherwise stated, AV2 metrics use $K=6$ and displacement errors are measured in meters.

For AV2 joint prediction, ADE and FDE are first averaged over the scored actors within each predicted world, and the world with the smallest average FDE is selected. We report its avgMinADE and avgMinFDE. ActorMR is the fraction of actors in the selected world whose final error exceeds 2~m, and avgBrier-minFDE adds the same confidence penalty to the selected world's avgMinFDE.

For WOMD, minADE and minFDE take the minimum displacement error over the $K$ predictions at each evaluation cutoff. A prediction is a match when its endpoint lies within the official lateral and longitudinal thresholds after rotating the error into the ground-truth agent frame. The base threshold pairs are $(1.0,2.0)$, $(1.8,3.6)$, and $(3.0,6.0)$ meters at 3, 5, and 8 seconds, respectively, and are scaled between 0.5 and 1.0 according to the agent's initial speed. MR is the fraction of agents for which no mode matches. Overlap rate (OR) measures how often the highest-confidence prediction overlaps another object visible at the prediction time at any step up to the cutoff.

Mean average precision (mAP) evaluates confidence ranking after grouping ground-truth trajectories into motion buckets. Matching predictions are true positives and misses are false positives, with at most one true positive per agent; precision--recall curves are computed per bucket and then averaged. Soft mAP differs only in its treatment of additional matching modes: after the highest-confidence match is counted as the true positive, further matches are ignored instead of counted as false positives. The main benchmark table reports the official mAP v2 and Soft mAP v2 values averaged over the three agent types and the 3, 5, and 8-second cutoffs; minFDE@8s is the corresponding all-object result at the 8-second cutoff. The horizon analysis reports minADE and minFDE at each cutoff separately using the same trained model.

\subsection{Training and Evaluation Protocol}

FutureLift receives frozen proposal coordinates and logits from each backbone and does not access backbone queries, exported embeddings, or intermediate decoder features. Because this interface consists only of trajectories and scores, no backbone-specific architectural adaptation is required, although FutureLift is trained separately for each setting. For marginal experiments, FutureLift replaces the native refinement stage of QCNet or DONUT. For joint experiments, it refines the final proposal sets produced by FJMP, Forecast-MAE, or DONUT. We run the trained comparison refiners from their official repositories using the authors' released hyperparameters; accordingly, SmartRefine also receives the backbone trajectory features required by its official implementation. The daggered SRefiner results are instead reproduced from the authors' AV2 evaluation with matched FJMP and Forecast-MAE backbones. While the SRefiner paper defines its initial feature as an encoding of proposal geometry and agent pose, its released AV2 implementation loads both proposal coordinates and a precomputed 134-dimensional per-mode tensor named \texttt{embeds\_final}; this tensor is projected to initialize refinement. Thus, these SRefiner rows neither use our bit-identical proposal files nor satisfy FutureLift's proposal-only input contract. Except for the explicitly daggered SRefiner rows, refiners compared within a backbone block use the same frozen proposal files, split, preprocessing, $K$, evaluator version, and random seeds.

Future and scene features have dimension $d=128$, and the default refiner uses two heterogeneous attention layers with eight heads. We use the same edge-enhanced Transformer operator in all experiments. Every concrete relation operator has a separate FFN with hidden dimension $4d=512$ and LeakyReLU activation. Dropout of $0.1$ is applied to the attention weights after softmax and to the hidden FFN activations. Source, target, and edge inputs are layer-normalized, as are the attention and FFN branch outputs before residual addition; \cref{appendix:sec:attention_operator} gives the exact update structure. We use $R=3$ temporal nodes for the six-second AV2 horizon and $R=4$ for the eight-second WOMD horizon. Map and social neighborhoods are radius-based, keeping the lifted graph sparse as the number of agents grows.

We optimize FutureLift with AdamW~\cite{adamw} using a total batch size of 64. Training lasts 40 epochs: the learning rate is linearly warmed up to $5\times10^{-4}$ during the first epoch and then decayed to zero with a cosine schedule over the remaining 39 epochs. We apply no data augmentation. For every AV2 and WOMD validation experiment trained in our study, we use seeds 42, 1337, and 2026 and report the arithmetic mean over the three runs. The same seeds are used for every comparison refiner that we train; the daggered SRefiner values are published results and are therefore excluded from this seed-averaging statement. Metric-wise standard deviations do not exceed $3\times10^{-3}$ across our validation runs, so we report the means in the main tables and summarize dispersion here rather than appending a deviation to every cell. For each run, we select the checkpoint with the lowest validation loss. This seed averaging applies only to validation-split results. For the official WOMD test submission, we randomly select one of the three trained seeds rather than choosing a model based on test performance.


\subsection{Relation-Specific Attention Operator}
\label{appendix:sec:attention_operator}

Each relation operator is a complete feature update with its own attention, gate, normalization, and FFN parameters. The self group first reads an agent's own history and then its other future intervals. Map operators read lane context followed by fine-grained map points. Social operators then read observed neighboring agents and, for joint proposals, coherent neighboring futures. Each operator sees the future-node features returned by its predecessor. Cross-mode attention is appended only in the final layer, followed in joint prediction by future-to-scene attention pooling over agents and intervals. Pooling has no edge-feature input.

For one relation $q$, source, target, and edge features are normalized before computing the queries, keys, and values in \cref*{eq:relation_attention}. Edge features enter both the key and value projections. Source and target normalizers are separate when their node types differ and shared for future-to-future operators. Let $\mathbf{m}_v$ concatenate all head outputs and let $\bar{\mathbf{h}}_v$ be the normalized target input. The attention update includes a channel-wise sigmoid gate,
\begin{align}
    \mathbf{r}_v^q &= [\mathbf{m}_v,\bar{\mathbf{h}}_v,\mathbf{m}_v-\bar{\mathbf{h}}_v], \\
    \mathbf{g}_v^q &= \sigma\!\left(G_q(\mathbf{r}_v^q)\right), \\
    \mathbf{z}_v^q &= (1-\mathbf{g}_v^q)\odot\mathbf{m}_v
        +\mathbf{g}_v^q\odot S_q(\bar{\mathbf{h}}_v),
    \label{appendix:eq:attention_gate}
\end{align}
where brackets denote concatenation, $G_q$ is an affine map, and $S_q$ projects the normalized target feature. Thus, the gate depends on the aggregated message, the target state, and their difference. An output projection $O_q$ and two normalized residual branches complete the operator:
\begin{align}
    \mathbf{u}_v &= \mathbf{h}_v+N_A^q\!\left(O_q(\mathbf{z}_v^q)\right), \\
    \mathbf{h}'_v &= \mathbf{u}_v+N_B^q\!\left(\operatorname{FFN}_q(N_F^q(\mathbf{u}_v))\right).
    \label{appendix:eq:relation_residual_update}
\end{align}
Here $N_A^q$, $N_B^q$, and $N_F^q$ are separate learned LayerNorm modules. In particular, $N_A^q$ and $N_B^q$ normalize branch outputs before addition, rather than normalizing the sum. The FFN consists of affine maps $d\to4d\to d$, with LeakyReLU and hidden dropout between them. The resulting $\mathbf{h}'_v$ immediately replaces the target state for the next relation. Separate relation operators are not combined by a shared FFN, and their sequential feature updates do not move graph anchors or decode intermediate coordinates.

\subsection{Efficiency Protocol}
\label{appendix:sec:efficiency_protocol}

Parameter counts and GFLOPs in \cref{appendix:tab:detailed_efficiency} cover the backbone and refiner combined. Latencies in \cref*{tab:temporal_resolution,appendix:tab:refinement_passes} cover only refinement; \cref{appendix:tab:detailed_efficiency} reports backbone and refinement latencies separately as backbone $\mid$ refinement. All three tables use an NVIDIA RTX 6000 Ada Generation GPU with PyTorch~2.5, FP32 precision, and a batch size of one. Refinement timing includes graph construction and the refiner forward pass, including FutureLift's scene encoder.

\begin{figure*}[t]
    \centering
    \begin{subfigure}[t]{0.34\textwidth}
        \centering
        \includegraphics[height=2.75cm]{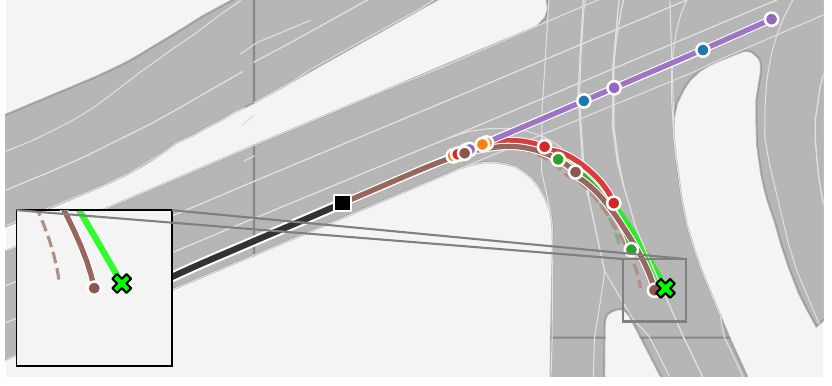}
        \caption{Lane-alignment correction.}
        \label{appendix:fig:res_1}
    \end{subfigure}
    \begin{subfigure}[t]{0.40\textwidth}
        \centering
        \includegraphics[height=2.75cm]{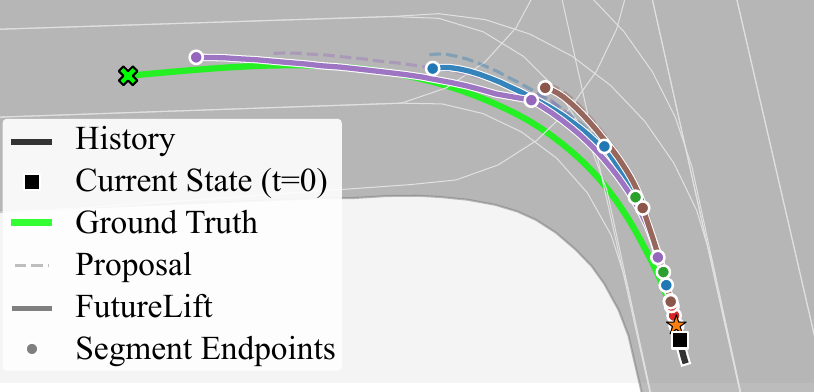}
        \caption{Correction along curved lane geometry.}
        \label{appendix:fig:res_2}
    \end{subfigure}
    \begin{subfigure}[t]{0.23\textwidth}
        \centering
        \includegraphics[height=2.75cm]{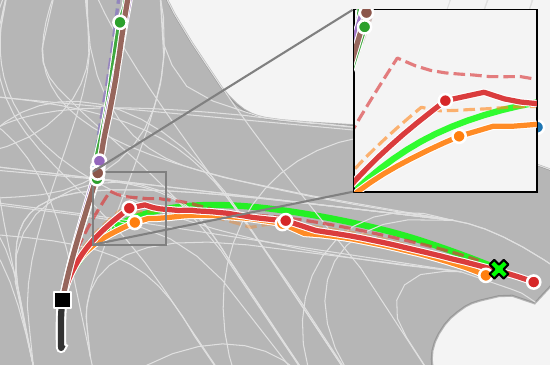}
        \caption{Correction of a local proposal discontinuity.}
        \label{appendix:fig:res_3}
    \end{subfigure}
    \caption{\textbf{Representative FutureLift corrections on the AV2 validation split.} Each panel compares a frozen backbone proposal with the corresponding FutureLift-refined hypothesis and ground truth. FutureLift corrects lateral drift, curved-lane misalignment, and a local trajectory discontinuity.}
    \label{appendix:fig:quali}
\end{figure*}

\subsection{Complete AV2 Benchmark Context}

The main tables prioritize controlled refinement comparisons. For completeness, \cref{appendix:tab:av2_marginal_context,appendix:tab:av2_joint_context} place the strongest FutureLift systems alongside additional published methods and include the official $K=1$ displacement metrics. For both AV2 test tasks, we report the FutureLift configuration selected on the validation split, which uses frozen DONUT proposals. Because the AV2 motion-forecasting leaderboard no longer accepts new submissions, we evaluate this configuration locally using the test annotations released by the dataset authors and the official evaluator. The resulting values are computed on the AV2 test split but therefore do not appear on the archived leaderboard. Published leaderboard values are included for benchmark context and are not used as evidence for identical-input refinement comparisons.

\subsection{Detailed System Cost}
\begin{table}[h]
    \centering
    \caption{Efficiency on an RTX 6000 Ada GPU under \cref{appendix:sec:efficiency_protocol}. Parameters and GFLOPs cover backbone and refiner combined; mean latencies are backbone $\mid$ refinement, in ms. Refinement includes graph construction. FMAE denotes Forecast-MAE.}
    \label{appendix:tab:detailed_efficiency}
    \scriptsize
    \setlength{\tabcolsep}{2.6pt}
    \begin{tabular}{@{}lllccc@{}}
        \toprule
        \shortstack{Dataset /\\task} & Backbone & Refiner & \shortstack{Params\\(M)} & GFLOPs & \shortstack{Mean latency (ms)\\Backb. $\mid$ refin.} \\
        \midrule
        AV2 marginal & QCNet & Native & 7.7 & 55.8 & $37 \mid 36$ \\
        AV2 marginal & QCNet & SmartRefine & 8.0 & 56.2 & $37 \mid 23$ \\
        AV2 marginal & QCNet & FutureLift & 13.8 & 84.4 & $37 \mid 37$ \\
        AV2 joint & FMAE & SRefiner & 3.5 & 27.8 & $24 \mid 23$ \\
        AV2 joint & FMAE & FutureLift & 10.1 & 65.2 & $24 \mid 41$ \\
        AV2 joint & FJMP & SRefiner & 5.9 & 45.6 & $40 \mid 18$ \\
        AV2 joint & FJMP & FutureLift & 13.1 & 78.0 & $40 \mid 36$ \\
        AV2 joint & DONUT & FutureLift & 12.2 & 94.6 & $82 \mid 42$ \\
        WOMD marginal & DONUT & Native & 9.0 & 76.2 & $107 \mid 36$ \\
        WOMD marginal & DONUT & FutureLift & 12.1 & 98.1 & $107 \mid 35$ \\
        \bottomrule
    \end{tabular}
\end{table}

\subsection{Number of Refinement Passes}
\label{appendix:sec:refinement_pass_ablation}

We evaluate whether repeatedly applying the refinement stage improves over FutureLift's default single-pass design using QCNet proposals on AV2 marginal prediction. Here, $J$ denotes the number of outer refinement passes and is distinct from the number of attention layers within one pass. For $J\geq1$, the winning mode used for supervision is selected after refinement, consistent with the training objective in \cref*{sec:method}; $J=0$ leaves the initial predictions unrefined.

\begin{table}[h]
    \centering
    \caption{Effect of the number of refinement passes $J$ on the AV2 marginal validation split. Displacement errors are in meters. Latency excludes the backbone and is measured on an RTX 6000 Ada GPU; $J=0$ incurs no refinement cost.}
    \label{appendix:tab:refinement_passes}
    \scriptsize
    \setlength{\tabcolsep}{2.7pt}
    \begin{tabular}{@{}lccccc@{}}
        \toprule
        $J$ & \shortstack{b-minFDE$_6$\\$\downarrow$} & \shortstack{minFDE$_6$\\$\downarrow$} & \shortstack{minADE$_6$\\$\downarrow$} & \shortstack{MR$_6$\\$\downarrow$} & \shortstack{Refinement\\latency (ms) $\downarrow$} \\
        \midrule
        0 & 1.875 & 1.235 & 0.644 & 0.162 & \textbf{0} \\
        \textbf{1} & \textbf{1.743} & \textbf{1.091} & \textbf{0.584} & \textbf{0.124} & 37 \\
        2 & 1.771 & 1.135 & 0.602 & 0.133 & 70 \\
        3 & 1.797 & 1.167 & 0.621 & 0.141 & 91 \\
        \bottomrule
    \end{tabular}
\end{table}

A single refinement pass provides the best accuracy, reducing minFDE by 11.7\% and minADE by 9.3\% relative to no refinement at a refinement latency of 37~ms. Two and three passes increase this cost to 70 and 91~ms, respectively, while consistently degrading the spatial metrics, supporting the single-pass design used throughout the paper.

\subsection{Geometric Diagnostic Protocol}
\label{appendix:subsec:geometric_diagnostics}

On the AV2 joint validation split with DONUT proposals, we compare all modes before and after refinement using matched valid observations. An eligible trajectory is off-road if any predicted center point lies outside the union of native AV2 drivable-area polygons; boundary points count as on-road. We exclude pedestrians and bicycle-related types and omit 28 scenes with unusable polygons. Among 549,678 eligible agent--mode trajectories, refinement resolves 979 off-road violations and introduces 124. Collision checks compare oriented agent footprints at matching future waypoints within the same joint mode, restricted to evaluated agents with valid dimensions and headings. A pair--mode combination counts as colliding if any eligible waypoint overlaps; touching footprints count as collisions. Across 1,501,938 eligible pair--mode combinations, the collision rate decreases slightly from 3.846\% to 3.353\%; the overlapping pair-timestep rate also decreases from 0.900\% to 0.722\%. These discrete open-loop checks do not measure collisions between waypoints or establish downstream planning benefit.

\subsection{Qualitative Analysis}
\label{appendix:subsec:qualitative}

\Cref{appendix:fig:quali} illustrates representative geometric corrections made by FutureLift on the AV2 validation split. Each scene compares the same trajectory hypothesis before and after refinement. In \cref{appendix:fig:res_1}, FutureLift reduces lateral drift from the local lane geometry. In \cref{appendix:fig:res_2}, it better follows a curved lane segment, while \cref{appendix:fig:res_3} shows the correction of a localized discontinuity in the initial proposal. These examples illustrate the local corrections enabled by proposal-conditioned future-graph reasoning; they do not by themselves establish physical feasibility or closed-loop safety.



\section{Detailed Node and Edge Descriptions}
The scene depicted in \cref{appendix:fig:all_nodes} is used in the following to visualize all different edges. The scene consists of a green and purple vehicle, each with 4 timesteps. There are 4 lanes in the scene. For better visibility the area of the lane is also colorized. Every lane holds multiple point nodes. Lastly the example scene holds agent-future nodes for 2 modes, each consisting of 3 segments.
\begin{figure}[h]
    \centering
    \includegraphics[width=\columnwidth]{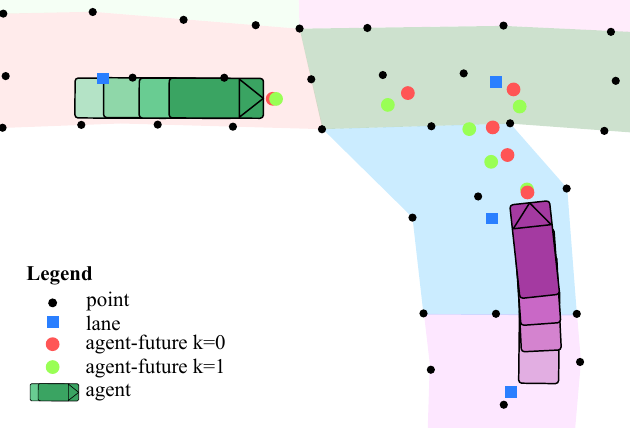}
    \caption{Example scene}
    \label{appendix:fig:all_nodes}
\end{figure}
\begin{figure}[h]
    \centering
    \includegraphics[width=\columnwidth]{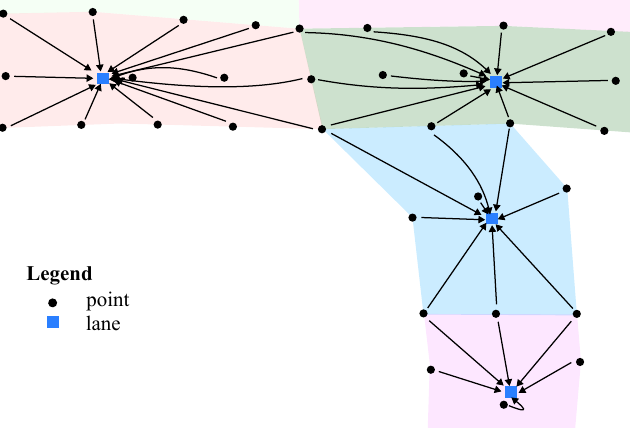}
    \caption{\textbf{(point, belongs-to, lane)}: All points belonging to the same lane, are connected to their respected lane node.}
    \label{appendix:fig:point_to_lane}
\end{figure}
\begin{figure}[h]
    \centering
    \includegraphics[width=\columnwidth]{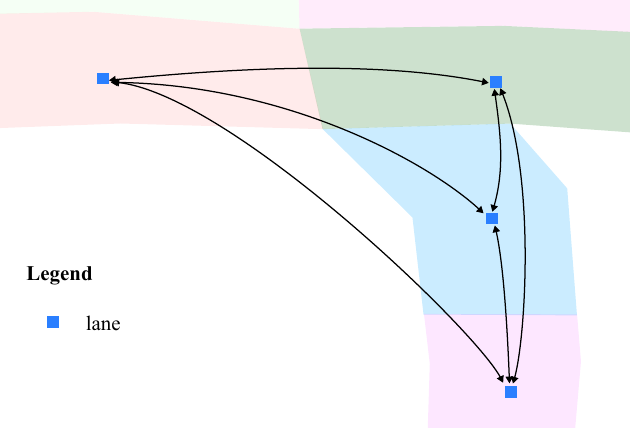}
    \caption{\textbf{(lane, to, lane)}: Lane nodes are connected to all neighbors within a radius of 150m.}
    \label{appendix:fig:lane_to_lane}
\end{figure}
\begin{figure}[h]
    \centering
    \includegraphics[width=\columnwidth]{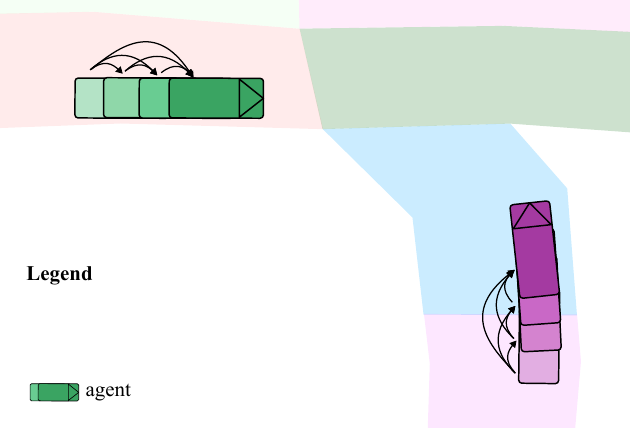}
    \caption{\textbf{(agent, self, agent)}: Agent nodes are connected to up to 10 subsequent neighbors.}
    \label{appendix:fig:agent_self_agent}
\end{figure}
\begin{figure}[h]
    \centering
    \includegraphics[width=\columnwidth]{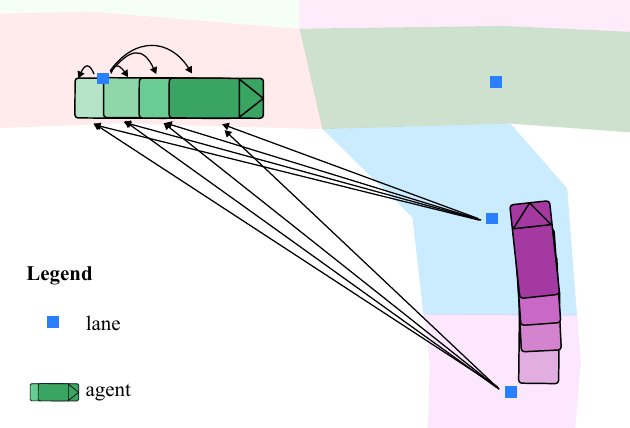}
    \caption{\textbf{(lane, gives-traffic-info, agent)}: Lanes are connected to all agents within a radius of 150m. For better visibility only edges to the green vehicle are depicted.}
    \label{appendix:fig:lane_to_agent}
\end{figure}
\begin{figure}[h]
    \centering
    \includegraphics[width=\columnwidth]{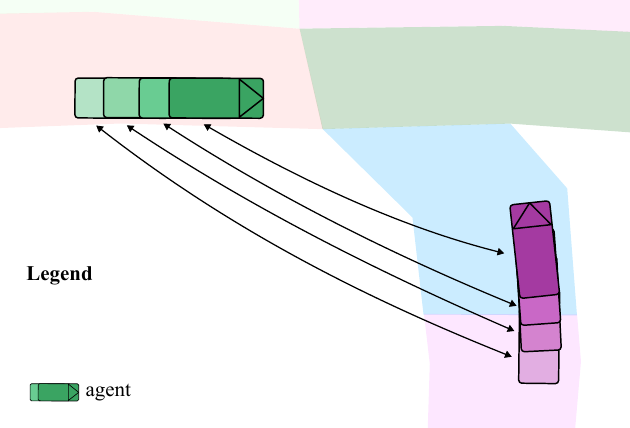}
    \caption{\textbf{(agent, social, agent)}: Agent nodes are connected to agent nodes from neighboring traffic participants at the concurrent timestamp within a 50m radius.}
    \label{appendix:fig:agent_social_agent}
\end{figure}
\begin{figure}[h]
    \centering
    \includegraphics[width=\columnwidth]{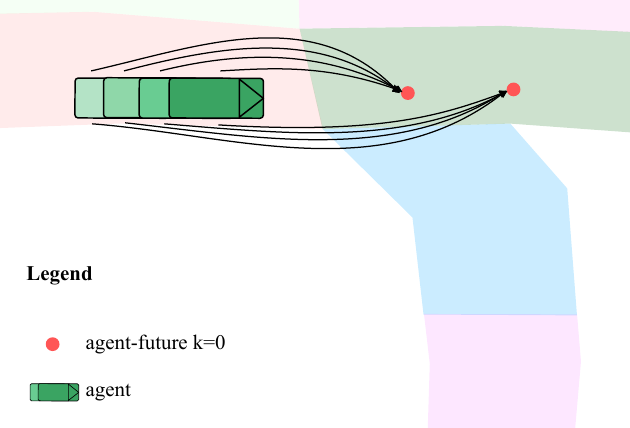}
    \caption{\textbf{(agent, self, agent-future)}: Agent-future nodes receive information from up to 30 previous agent nodes.}
    \label{appendix:fig:agent_self_agentfuture}
\end{figure}
\begin{figure}[h]
    \centering
    \begin{subfigure}{\columnwidth}
        \includegraphics[width=\columnwidth]{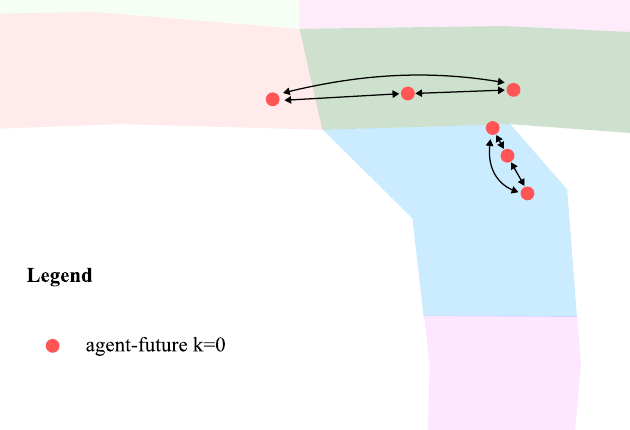}
    \end{subfigure}
    \caption{\textbf{(agent-future, self, agent-future)}: Agent-future nodes receive updates from other agent-future nodes belonging to the same trajectory, i.e. the same mode. For greater visibility only nodes for k = 0 are depicted.}
    \label{appendix:fig:agentfuture_self_agentfuture}
\end{figure}
\begin{figure}[h]
    \centering
    \includegraphics[width=\columnwidth]{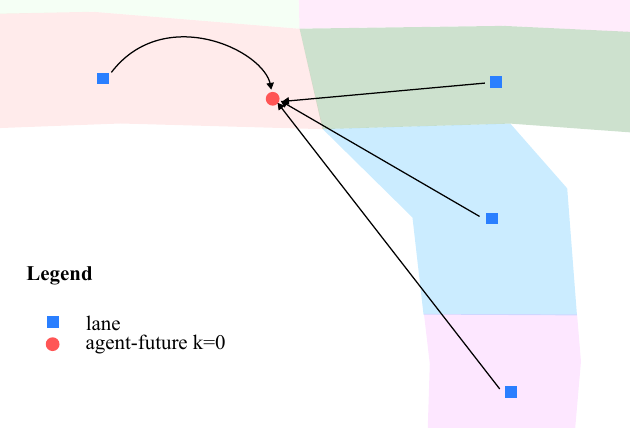}
    \caption{\textbf{(lane, gives-traffic-info, agent-future)}: Lane nodes are connected to agent future nodes within a radius of 150m.}
    \label{appendix:fig:lane_to_agentfuture}
\end{figure}
\begin{figure}[h]
    \centering
    \includegraphics[width=\columnwidth]{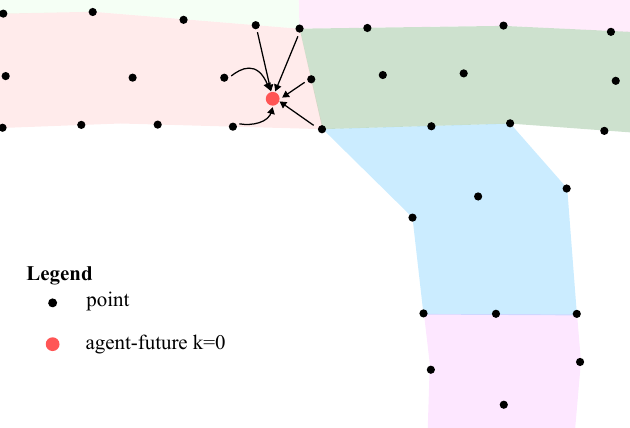}
    \caption{\textbf{(point, gives-traffic-info, agent-future)}: Point nodes are connected to agent-future nodes within a radius of 10m.}
    \label{appendix:fig:point_to_agentfuture}
\end{figure}
\begin{figure}[h]
    \centering
    \includegraphics[width=\columnwidth]{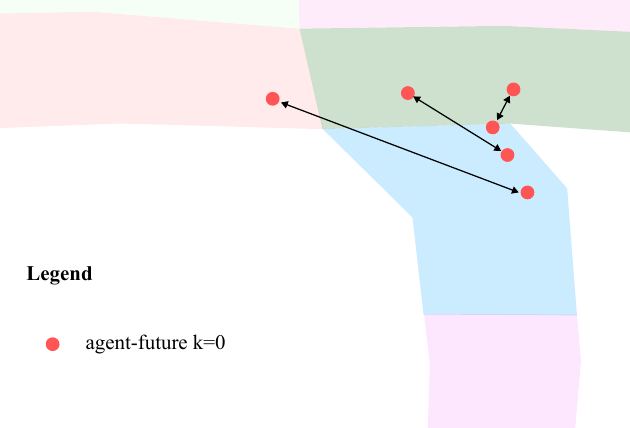}
    \caption{\textbf{(agent-future, social, agent-future)}: Agent-future nodes are connected to neighboring traffic participants in the same joint mode and future interval within a 150m radius. Only k = 0 is depicted for simplicity.}
    \label{appendix:fig:agentfuture_social_agentfuture}
\end{figure}
\begin{figure}[h]
    \centering
    \includegraphics[width=\columnwidth]{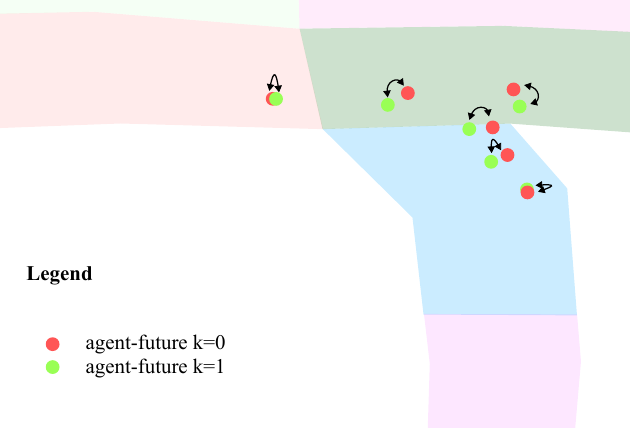}
    \caption{\textbf{(agent-future, mode, agent-future)}: Agent-future nodes belonging to the same future interval and traffic participant are connected across different modes.}
    \label{appendix:fig:agentfuture_mode_agentfuture}
\end{figure}

\end{document}